\documentclass[conference]{IEEEtran}
\IEEEoverridecommandlockouts
\usepackage{cite}
\usepackage{amsmath,amssymb,amsfonts}
\usepackage{algorithmic}
\usepackage{graphicx}
\usepackage{textcomp}
\usepackage{xcolor}

\usepackage{caption}
\usepackage{booktabs}
\usepackage{multirow}  
\usepackage{makecell}  
\usepackage{booktabs}  

\usepackage{url}
\usepackage[utf8]{inputenc}
\DeclareUnicodeCharacter{FFFD}{}

\def\BibTeX{{\rm B\kern-.05em{\sc i\kern-.025em b}\kern-.08em
    T\kern-.1667em\lower.7ex\hbox{E}\kern-.125emX}}
\begin{document}

\title{Causal GNNs: A GNN-Driven Instrumental Variable Approach for Causal Inference in Networks}

\author{\IEEEauthorblockN{1\textsuperscript{st} Xiaojing Du}
\IEEEauthorblockA{\textit{STEM} \\
\textit{University of South Australia}\\
Adelaide, Australia \\
xiaojing.du@mymail.unisa.edu.au}
\and
\IEEEauthorblockN{2\textsuperscript{nd} Feiyu Yang\thanks{Corresponding author: Feiyu Yang, feiyu@my.swjtu.edu.cn}}
\IEEEauthorblockA{\textit{School of Information Science and Technology} \\
\textit{Southwest Jiaotong University}\\
Chengdu, China \\
feiyu@my.swjtu.edu.cn}
\and
\IEEEauthorblockN{3\textsuperscript{nd} Wentao Gao}
\IEEEauthorblockA{\textit{STEM} \\
\textit{University of South Australia}\\
Adelaide, Australia \\
wentao.gao@mymail.unisa.edu.au}
\and
\IEEEauthorblockN{4\textsuperscript{nd} Xiongren Chen}
\IEEEauthorblockA{\textit{STEM} \\
\textit{University of South Australia}\\
Adelaide, Australia \\
xiongren.chen@mymail.unisa.edu.au}
}

\maketitle

\begin{abstract}

As network data applications continue to expand, causal inference within networks has garnered increasing attention. However, hidden confounders complicate the estimation of causal effects. Most methods rely on the strong ignorability assumption, which presumes the absence of hidden confounders—an assumption that is both difficult to validate and often unrealistic in practice. To address this issue, we propose CgNN, a novel approach that leverages network structure as instrumental variables (IVs), combined with graph neural networks (GNNs) and attention mechanisms, to mitigate hidden confounder bias and improve causal effect estimation. By utilizing network structure as IVs, we reduce confounder bias while preserving the correlation with treatment. Our integration of attention mechanisms enhances robustness and improves the identification of important nodes. Validated on two real-world datasets, our results demonstrate that CgNN effectively mitigates hidden confounder bias and offers a robust, GNN-driven IV framework for causal inference in complex network data.

\end{abstract}


\section{Introduction}
\begin{figure}[t]
	\centering
	\includegraphics[scale=0.27]{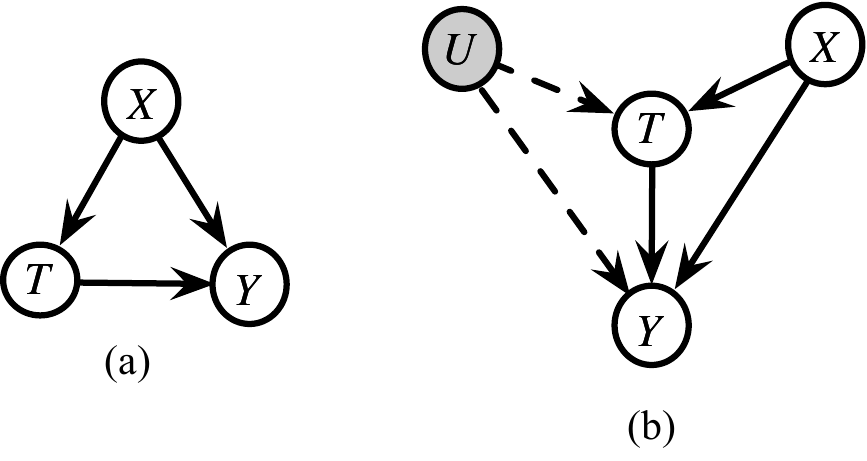}
	\caption{Causal DAGs illustrating challenges in causal effect estimation within networks. \(\mathbf{X}\) and \(\mathbf{U}\) are observed features and hidden confounders, \(\mathbf{T}\) and \(\mathbf{Y}\) represent treatment and outcome. (a) assumes strong ignorability, (b) includes hidden confounder \(\mathbf{U}\).}
	\label{fig:1}
\end{figure}

Causal inference is essential in fields such as epidemiology~\cite{rothman2005causation}, economics~\cite{varian2016causal}, and medicine~\cite{yazdani2015causal}. Estimating causal effects in network data is often complicated by confounding factors, especially hidden confounders, which introduce significant challenges. When hidden confounders influence both the treatment and the outcome, they can create spurious associations, leading to inaccurate causal effect estimates~\cite{vanderweele2008causal,gao2024deconfounding}, as illustrated in the causal directed acyclic graph (DAG)~\cite{pearl2009causality} in Fig.~\ref{fig:1}. In such scenarios, traditional methods frequently fail to properly identify causal effects~\cite{pearl2009causality}.


Most existing causal inference methods for network data \cite{zhang2021treatment,jiang2023cf,forastiere2021identification,du2024estimating} rely on the \textit{strong ignorability assumption}. This assumption posits that, given covariates (i.e., features), the treatment assignment is independent of the potential outcomes. Consequently, the estimation of treatment effects is not affected by hidden confounders. In other words, this assumption requires that all confounders influencing the treatment effect be fully observable. However, in real-world applications, identifying all potential confounders is often unrealistic, making it challenging to maintain the strong ignorability assumption in practice.

The instrumental variable (IV) method is an effective approach for identifying causal effects in the presence of hidden confounders in independent and identically distributed (i.i.d.) data~\cite{cheng2024instrumental}. An IV is an exogenous variable that is related to the treatment but does not directly affect the outcome. IV-based methods typically follow a two-stage process: first, the IV is used to estimate the treatment, then the estimated treatment is used to predict the outcome. The two-stage least squares (2SLS) method\cite{angrist1995two} is a widely used IV approach that applies a linear model to estimate treatment effects.

Peer interference~\cite{bramoulle2009identification} is common in network data because individuals are interconnected and can influence each other's outcomes. This means that one person's treatment can affect the outcomes of others~\cite{schwartz2012extending}. For example, in epidemiology, as shown in Fig~\ref{fig:2}, we want to estimate the causal effect of vaccination on individual infection status. Suppose we have a social network where each node represents an individual, and edges represent social relationships between individuals. \(\mathbf{X}_i\) represents individual \(i\)'s features (e.g., health conditions). The variable \(T_i\) indicates whether individual \(i\) has been vaccinated, and \(Y_i\) represents their health status. \(Y_i\) can also be influenced by the treatment \(T_j\) of peers (details are discussed in the preliminary section). Hidden confounders \( \mathbf{U}_i \), such as lifestyle or socioeconomic status, can affect both the decision to get vaccinated and the subsequent infection status, leading to biased estimates of the vaccination effect.

The topology of networks is ubiquitous in various types of observational data, such as patient social networks and disease transmission networks. When certain confounding factors are difficult to measure directly, we can attempt to capture their patterns using the network structuree\cite{ma2023look}. To address these hidden confounding factors, the network structure can be employed as an IV.




In summary, by leveraging network structure information, we can effectively control for hidden confounding factors within networks. Moreover, by integrating GNNs~\cite{chen2019equivalence} with attention mechanisms~\cite{niu2021review}, we can capture the complex dependencies between nodes and accurately distinguish the influence of different peer (i.e., neighboring) nodes on the target node. The key contributions of our work are outlined as follows:

\begin{itemize}
    \item \textbf{Problem.} We propose a GNN-based approach that integrates causal inference with IVs to address hidden confounders in network data, especially in the presence of peer interference.
    
    \item \textbf{Method.} We introduce CgNN, a novel model that effectively distinguishes peer influences to better capture complex dependencies between nodes.
    
    \item \textbf{Experiments.} We conduct extensive experiments on real-world datasets, demonstrating the ability of the CgNN approach to effectively handle hidden confounders.
\end{itemize}

\section{Preliminary}
\label{section:preliminary}

This section outlines the primary notations and problem setup. Variables are represented by uppercase letters, while their corresponding values are shown in lowercase. Bold uppercase letters denote vectors or matrices, and bold lowercase letters represent their respective values.

We define the observational data as ${\mathbf{V}, \mathbf{X}, \mathbf{A}, \mathbf{T}, \mathbf{Y}}$, where $\mathbf{V}$ represents nodes (e.g., individuals), $\mathbf{X}$ refers to node features (e.g., health conditions), $\mathbf{A}$ is the adjacency matrix (e.g., network structure), $\mathbf{T}$ denotes treatments (e.g., vaccination), and $\mathbf{Y}$ indicates outcomes (e.g., infection status). The treatment $t_i \in {0, 1}$ is binary, with $t_i = 1$ indicating treatment.

\begin{figure}[t]
	\centering
	\includegraphics[scale=0.32]{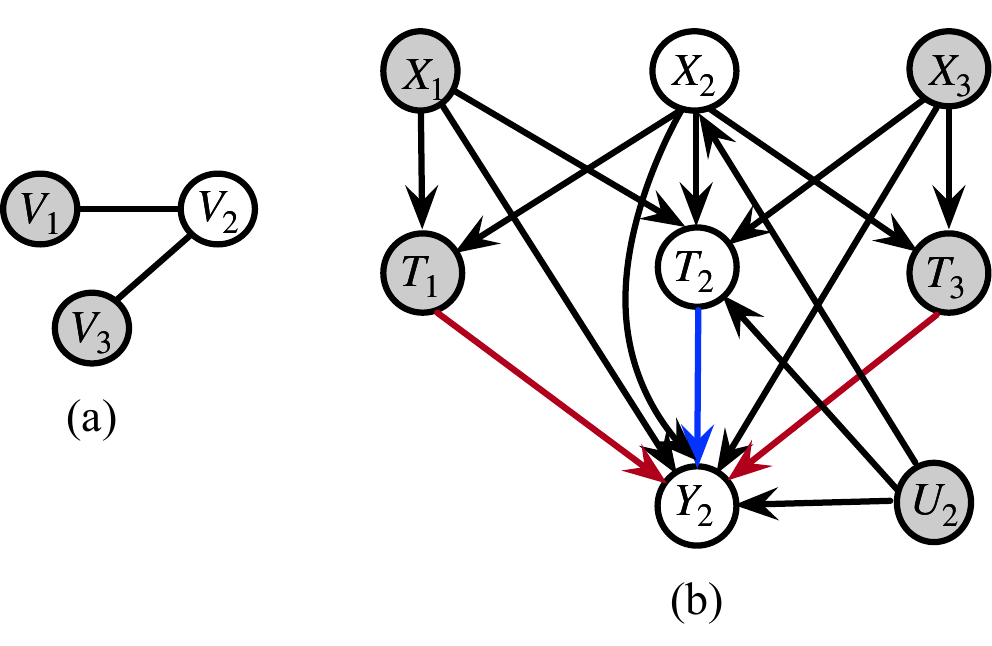}
	\caption{(a) illustrates the network structure. (b) focuses on node 2, which has nodes 1 and 3 as its peers within the network.}
	\label{fig:2}
\end{figure}

The outcome \(Y_i\) for unit \(i\) is determined by its own features \(\mathbf{X}_i\), treatment \(T_i\), hidden confounders \(\mathbf{U}_i\), and the features \(\{\mathbf{X}_j\}_{j \in \mathcal{N}_i}\) and treatments \(\{T_j\}_{j \in \mathcal{N}_i}\) of its peers, where \(j \in \mathcal{N}_i\) represents the first-order neighboring nodes, as illustrated in Fig~\ref{fig:2}. To account for the varying interference from neighbors' treatments, we define \(z_i = \sum_{j \in \mathcal{N}_i} w_{ij} \cdot t_j\), where \(z_i\) captures the influence of peer treatments on unit \(i\).

\begin{equation}
w_{ij} = \frac{1}{1 + \sum_{k} P_i(k) \log \left( \frac{P_j(k)}{P_i(k)} \right)}
\label{eq:wij}
\end{equation}

where $P_i(k)$ and $P_j(k)$ are the probability distributions for features of nodes $i$ and $j$ over the $k$-th dimension.

\textbf{Problem Definition.} Based on the network data $\{\mathbf{V}, \mathbf{X}, \mathbf{A}, \mathbf{T}, \mathbf{Y}\}$, our objective is to accurately estimate the \textit{Main Effects} (ME) (i.e., the causal effect of $T_i$ on $Y_i$ via $T_i \rightarrow Y_i$, shown by the blue line in Fig.~\ref{fig:2}b), \textit{Peer Effects} (PE) (i.e., the causal effect of $T_j$ on $Y_i$ via $T_j \rightarrow Y_i$, shown by the red line), and \textit{Total Effects} (TE), which captures both effects.

We do not assume Strong Ignorability, allowing for hidden confounders and addressing the biases they introduce.

\textbf{Assumption 1} (Strong Ignorability) Given the features \( \mathbf{X}_i \) and the peers' features \( \{\mathbf{X}_j\}_{j \in \mathcal{N}_i} \), the potential outcome \( Y(t_i, z_i) \) does not depend on the treatment \( T_i \) and peer influence \( Z_i \), i.e.,
\( Y(t_i, z_i) \perp\!\!\!\perp T_i, Z_i \mid \mathbf{X}_i, \{\mathbf{X}_j\}_{j \in \mathcal{N}_i} \).

\section{The Proposed CgNN Method}

This section outlines our study's objectives, the key assumptions to address hidden confounders, and the methods.

\subsection{Estimation Targets and Assumptions}

Our target estimands focus on three effects~\cite{jiang2022estimating,chen2024doubly}:

\paragraph{Main effects} 
\[
E\left( Y_i(t_i, 0) - Y_i(t'_i, 0) \mid \mathbf{X}_i, \{\mathbf{X}_j\}_{j \in \mathcal{N}_i} \right),
\]
the effect of an individual’s own treatment \(t_i\) on their outcome.

\paragraph{Peer effects} 
\[
E\left( Y_i(0, z_i) - Y_i(0, z'_i) \mid \mathbf{X}_i, \{\mathbf{X}_j\}_{j \in \mathcal{N}_i} \right),
\]
the effect of the treatment of an individual’s peers \(z_i\) on their outcome.

\paragraph{Total effects}
\[
E\left( Y_i(t_i, z_i) - Y_i(t'_i, z'_i) \mid \mathbf{X}_i, \{\mathbf{X}_j\}_{j \in \mathcal{N}_i} \right),
\]
the joint effect of individual treatment \(t_i\) and peer influence \(z_i\) on the outcome.

We rely on the following assumptions to leverage the graph structure as an IV for estimating causal treatment effects.

\textbf{Assumption 2} (Relevance) The treatment \(\mathbf{T}\) is associated with the graph structure \(\mathbf{A}\), meaning that \(\mathbf{A}\) and \(\mathbf{T}\) are conditionally dependent given \(\mathbf{X}\), i.e., \(\mathbf{A} \not\perp\!\!\!\perp \mathbf{T} \mid \mathbf{X}\).

\textbf{Assumption 3} (Exclusion Restriction) The effect of \(\mathbf{A}\) on the outcome \(\mathbf{Y}\) is fully mediated by \(\mathbf{T}\), implying that changes in \(\mathbf{A}\) do not directly affect \(\mathbf{Y}\), i.e., \(\mathbf{Y}(\mathbf{T}, \mathbf{A}) = \mathbf{Y}(\mathbf{T}, \mathbf{A}')\) for all \(\mathbf{A} \neq \mathbf{A}'\).

\textbf{Assumption 4} (Instrumental Unconfoundedness) There are no unblocked backdoor paths~\cite{pearl2009causality} from \(\mathbf{A}\) to \(\mathbf{Y}\), i.e., \(\mathbf{A} \perp\!\!\!\perp \mathbf{Y} \mid \mathbf{T},\mathbf{X}\).

These assumptions follow recent IV research~\cite{ma2023look} to ensure the graph structure is a valid IV for estimating effects.


\subsection{Implementation}

To estimate the ME, PE, and TE in networks, we adopt a two-stage IV approach driven by GNNs, as shown in Fig.~\ref{fig:3}. In the first stage, GNNs predict the treatment variable \(T\), eliminating hidden confounder bias. In the second stage, the predicted \(T\) is used to estimate the outcome \(Y\), allowing for more precise causal effect estimation. The attention mechanism assigns varying weights to each neighbor to capture their influence on the target node, calculated as follows:



\begin{figure}[t]
	\centering
	\includegraphics[scale=0.24]{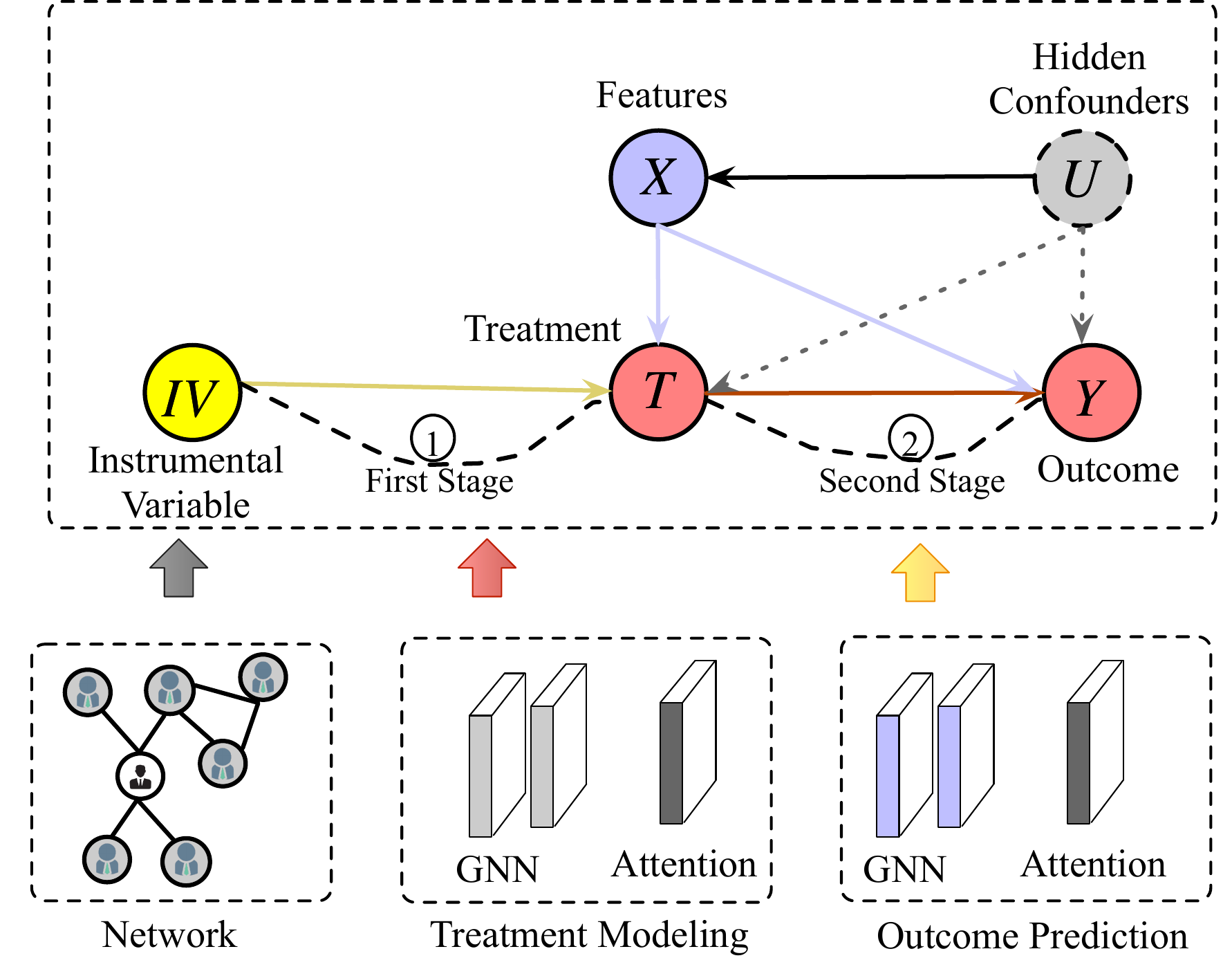}
	\caption{The workflow of our CgNN method for estimating ME, PE and TE within network data.}
	\label{fig:3}
\end{figure}

\begin{equation}
e_{ij} = \text{LeakyReLU}\left(\mathbf{A}^T \left[\mathbf{W} \mathbf{H}_i \| \mathbf{W} \mathbf{H}_j\right]\right)
\end{equation}

where \(e_{ij}\) is the attention score between node \(i\) and its neighbor \(j\), \(\mathbf{H}_i\) and \(\mathbf{H}_j\) are the feature representations of the nodes, \(\mathbf{W}\) represents the weight matrix, \(\mathbf{A}\) is the learnable attention vector, and \(\|\) denotes the concatenation of vectors.

Attention scores are normalized via softmax:

\begin{equation}
\alpha_{ij} = \frac{\exp(e_{ij})}{\sum_{k \in \mathcal{N}(i)} \exp(e_{ik})}
\end{equation}

where \(\alpha_{ij}\) represents the influence of neighbor \(j\) on node \(i\).

The loss functions for predicting the treatment and outcome are defined as follows:

\begin{equation}
\mathcal{L}_{T} = \frac{1}{N} \sum_{i=1}^{N} (T_i - \hat{T}_i)^2 + \lambda \|\mathbf{W}_T\|_2^2
\end{equation}
\begin{equation}
\mathcal{L}_{Y} = \frac{1}{N} \sum_{i=1}^{N} (Y_i - \hat{Y}_i)^2 + \lambda \|\mathbf{W}_Y\|_2^2
\end{equation}

Where \(T_i\) and \(Y_i\) are the observed treatment and outcome, \(\hat{T}_i\) and \(\hat{Y}_i\) are the predictions, and \(N\) is the number of nodes. \( \|\mathbf{W}_T\|_2^2 \) and \( \|\mathbf{W}_Y\|_2^2 \) are the \( L_2 \) norms~\cite{wu1996induced} of the weight matrices, and \( \lambda \) represents the regularization parameters.

\section{Experiment}

\subsection{Datasets}

We follow~\cite{jiang2022estimating,chen2024doubly,guo2020learning} in using semisynthetic datasets to evaluate our method. Specifically, we employ two publicly available datasets, BlogCatalog\footnote{\url{https://www.blogcatalog.com/}} (BC) and Flickr\footnote{\url{https://www.flickr.com/}}. In BlogCatalog, nodes represent bloggers, with edges denoting social connections and features extracted from profile keywords via a bag-of-words model. In Flickr, nodes represent users, with edges indicating friendships, and features derived from tags that users assign to their posts, reflecting their interests. Dataset details are presented in Table~\ref{table:datasets}.


\begin{table}[ht]
\centering
\caption{Datasets}
\begin{tabular}{lrr}
\toprule
& BlogCatalog & Flickr \\
\midrule
\# of Users & 5,196 & 7,575 \\
\# of Features & 8,189 & 12,047 \\
\# of Links & 171,743 & 239,738 \\
\bottomrule
\end{tabular}
\label{table:datasets}
\end{table}



\subsection{Simulation}

We simulate different variables based on the causal DAG in Fig.~\ref{fig:2}b as follows.

\begin{table*}[t]
\centering
\caption{The results show the $\epsilon_{PEHE}$ errors for causal effect estimation, with the best results highlighted in bold.}
\scriptsize 
\begin{tabular}{@{}clcccccc@{}} 
\toprule
Dataset & Effects & CFR(+N) & ND(+N) & TARNET(+N) & NetEst & TNet & CgNN \\ \midrule
\multirow{3}{*}{\thead{BC \\ (within-sample)}} 
& Main & 0.3195±0.0299 & 0.3488±0.0249 & 0.2830±0.0229 & 0.2390±0.0155 & 0.2257±0.0171 & \textbf{0.2174±0.0135} \\
& Peer & 0.2875±0.0151 & 0.3052±0.0135 & 0.2607±0.0140 & 0.2439±0.0138 & 0.2334±0.0127 & \textbf{0.1856±0.0098} \\
& Total & 0.1987±0.0112 & 0.2184±0.0124 & 0.1851±0.0136 & 0.1657±0.0102 & 0.1548±0.0114 & \textbf{0.1502±0.0059} \\ \cmidrule(lr){1-8}

\multirow{3}{*}{\thead{BC \\ (out-of-sample)}} 
& Main & 0.3184±0.0259 & 0.3488±0.0250 & 0.2898±0.0263 & 0.2441±0.0174 & 0.2323±0.0165 & \textbf{0.2263±0.0091} \\
& Peer & 0.2924±0.0163 & 0.3136±0.0147 & 0.2710±0.0131 & 0.2470±0.0141 & 0.2359±0.0134 & \textbf{0.1938±0.0123} \\
& Total & 0.2089±0.0124 & 0.2300±0.0117 & 0.1955±0.0140 & 0.1679±0.0110 & 0.1563±0.0121 & \textbf{0.1789±0.0045} \\ \cmidrule(lr){1-8}

\multirow{3}{*}{\thead{Flickr \\ (within-sample)}} 
& Main & 0.2575±0.0741 & 0.3011±0.0651 & 0.2215±0.0585 & 0.2397±0.0148 & 0.2285±0.0134 & \textbf{0.2075±0.0157} \\
& Peer & 0.2703±0.0182 & 0.2881±0.0167 & 0.2456±0.0174 & 0.2401±0.0162 & 0.2302±0.0151 & \textbf{0.1943±0.0104} \\
& Total & 0.1854±0.0153 & 0.2047±0.0139 & 0.1732±0.0148 & 0.1684±0.0127 & 0.1596±0.0113 & \textbf{0.1673±0.0091} \\ \cmidrule(lr){1-8}

\multirow{3}{*}{\thead{Flickr \\ (out-of-sample)}} 
& Main & 0.2646±0.0732 & 0.3112±0.0501 & 0.2237±0.0588 & 0.2421±0.0187 & 0.2304±0.0168 & \textbf{0.2184±0.0196} \\
& Peer & 0.2782±0.0169 & 0.2987±0.0173 & 0.2550±0.0160 & 0.2443±0.0174 & 0.2342±0.0158 & \textbf{0.1773±0.0120} \\
& Total & 0.1975±0.0140 & 0.2208±0.0144 & 0.1816±0.0154 & 0.1697±0.0138 & 0.1612±0.0124 & \textbf{0.1587±0.0088} \\ \bottomrule
\end{tabular}
\label{table1}
\end{table*}

\textbf{Hidden confounders.} The hidden confounders are generated as follows:

\begin{equation}
\mathbf{U}_i \sim \mathcal{N}(0, \mu \mathbf{I})
\label{eq:hidden_confounder}
\end{equation}

where \(\mathbf{I}\) is the identity matrix with dimensionality \(d_u\), representing the size of the hidden confounders. For our experiments, we set \(\mu = 20\) and \(d_u = 10\).

\textbf{Feature.} The node features are generated as follows:

\begin{equation}
\mathbf{X}_i = \mathbf{x}_i + \psi \mathbf{U}_i + \epsilon_x
\label{eq:feature_with_confounder_and_noise}
\end{equation}

where \( \mathbf{x}_i \) represents the observed node features, and \( \psi(\mathbf{U}_i) \) is a linear mapping from hidden confounders \(\mathbf{U}_i\) in \(\mathbb{R}^{d_u}\) to \(\mathbb{R}^{d_x}\). \(\epsilon_x\) is Gaussian noise.

\textbf{Treatment.} The treatment \(T_i\) is generated using the following equation, with the definition of \( w_{ij} \) detailed in the preliminary section:

\begin{equation}
\begin{aligned}
 & p(T_i = 1 \mid \mathbf{X}_i, \{\mathbf{X}_j\}_{j \in \mathcal{N}_i}, \mathbf{U}_i)\\ & = \sigma ( \alpha_0 \mathbf{w}_0 \mathbf{X}_i + \alpha_1 \sum_{j \in \mathcal{N}_i} w_{ij} \mathbf{w}_1 \mathbf{X}_j + \alpha_2 \mathbf{w}_2 \mathbf{U}_i +\epsilon_t )
\end{aligned}
\label{eq:t1_mod}
\end{equation}

where \( \mathbf{w}_0 \), \( \mathbf{w}_1 \), and \( \mathbf{w}_2 \) are randomly generated weight vectors. We set \(\alpha_0 = 1\), \(\alpha_1 = 0.5\), and \(\alpha_2 = 0.1\). \(\epsilon_t\) is a Gaussian noise term drawn from \(\mathcal{N}(0, 0.01^2)\).

The treatment is sampled from a Bernoulli distribution~\cite{chen1997statistical}:

\begin{equation}
T_i \sim \text{Bernoulli} \left( p(T_i = 1 \mid \mathbf{X}_i, \{\mathbf{X}_j\}_{j \in \mathcal{N}_i}), \mathbf{U}_i \right)
\label{eq:t2}
\end{equation}

\textbf{Potential Outcome.} The potential outcome \(Y_i\) is generated as follows:

\begin{equation}
\begin{aligned}
&p(Y_i \mid \mathbf{X}_i, T_i, \{\mathbf{X}_j\}_{j \in \mathcal{N}_i}, \{T_j\}_{j \in \mathcal{N}_i}, \mathbf{U}_i) \\
&= \sigma (\beta_0 \mathbf{w}_3 \mathbf{X}_i) + \sigma ( \beta_1 \sum_{j \in \mathcal{N}_i} w_{ij} \mathbf{w}_4 \mathbf{X}_j ) + \beta_2 T_i \\
&\quad + \beta_3 \sum_{j \in \mathcal{N}_i} w_{ij} T_j + \beta_4 \mathbf{w}_5 \mathbf{U}_i + \epsilon_y
\end{aligned}
\label{eq:simuy}
\end{equation}

where \(\epsilon_y \sim \mathcal{N}(0, 0.1^2)\) is a noise, \( \beta_k \sim \mathcal{U}(0, 1) \) for \( k = 0, 1, 2, 3, 4 \), \( \mathbf{w}_3 \), \( \mathbf{w}_4 \), \( \mathbf{w}_5 \) as randomly generated weight vectors.

\textbf{Baselines.} We evaluated our model against five baselines: (1) CFR~\cite{shalit2017estimating}, which uses integral probability metric (IPM)~\cite{sriperumbudur2012empirical} to balance distributions on independent and identically distributed data; (2) TARNet~\cite{shalit2017estimating}, a variant of CFR that omits the IPM; (3) NetDeconf~\cite{guo2020learning}, an extension of CFR designed for networks, using GNNs to address confounding variables. The models CFR+(N), TARNet+(N), and NetDeconf+(N) are further extended to account for peer effects; (4) NetEst~\cite{jiang2022estimating}, which incorporates adversarial learning~\cite{lowd2005adversarial} to estimate causal effects in network data; (5) TNet~\cite{chen2024doubly}, which employs target learning to enhance causal inference.


\textbf{Metrics.} We estimate model performance using Mean Squared Error (MSE)~\cite{wang2009mean} and Precision in Estimating Heterogeneous Effects (PEHE)~\cite{alaa2018limits}. MSE, defined as \(\epsilon_{MSE} = \frac{1}{m} \sum_{i=1}^{m} (\hat{Y}_i - Y_i)^2\)
measures the accuracy of counterfactual predictions. PEHE evaluates the precision in estimating causal effects, defined as
\(\epsilon_{PEHE} = \sqrt{\frac{1}{m} \sum_{i=1}^{m} \left[(\hat{Y}_i(t') - \hat{Y}_i(t)) - (Y_i(t') - Y_i(t))\right]^2}\). \(\hat{Y}_i\) and \(Y_i\) are predicted and ground truth outcomes, respectively. Lower values indicate better performance.

\begin{figure}[t]
	\centering
	\includegraphics[scale=0.177]{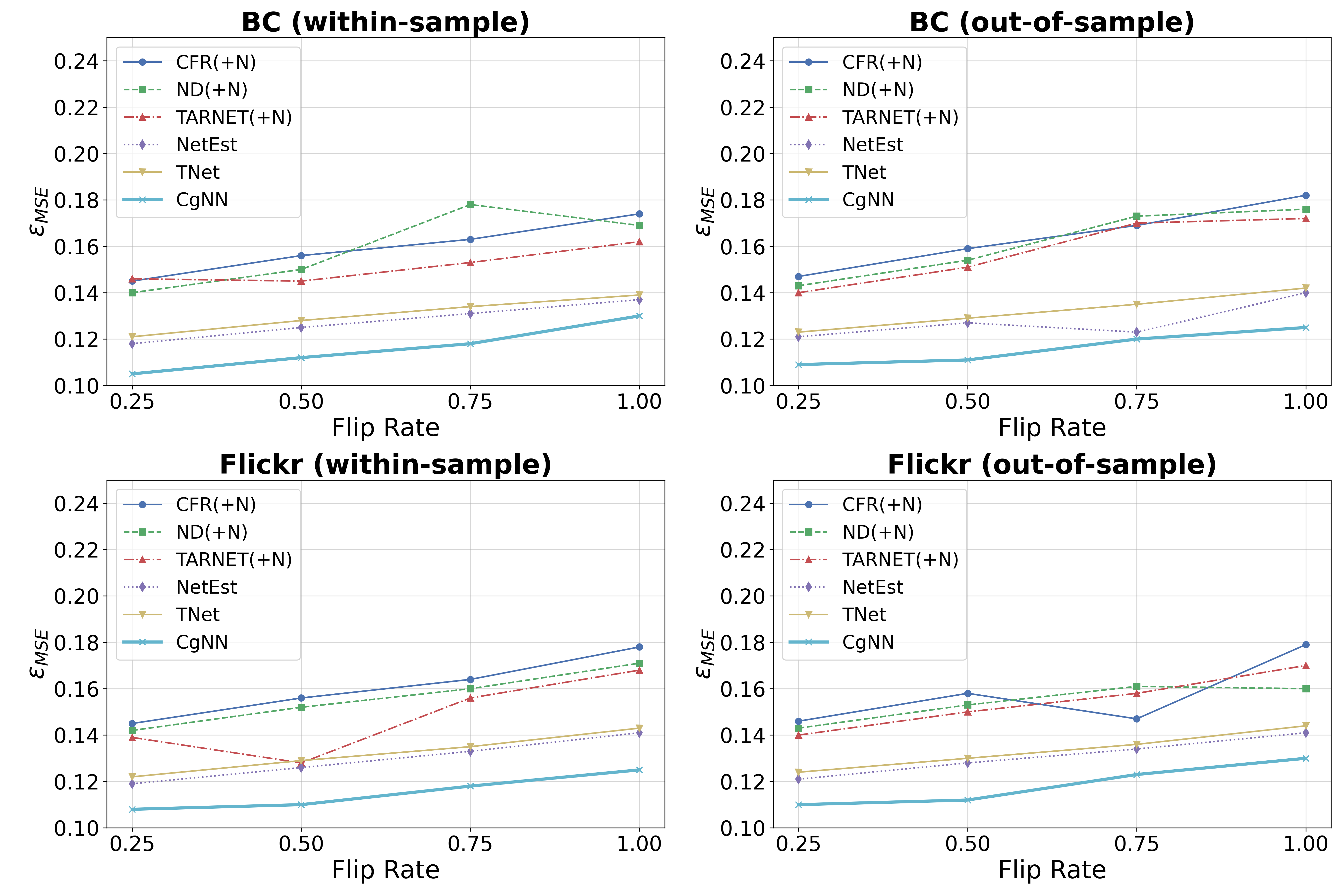}
	\caption{The results demonstrate how the counterfactual estimation error ($\epsilon_{MSE}$) correlates with the proportion of units experiencing treatment flips.}
	\label{fig:4}
\end{figure}

\textbf{Results.} We use the CgNN model to estimate ME, PE, and TE, evaluating both ``within-sample'' performance on the training network and ``out-of-sample'' generalization on the test network. The process is repeated 5 times, with the average and standard deviation reported. Table~\ref{table1} presents the $\epsilon_{PEHE}$ results for the BC and Flickr datasets. CgNN consistently outperforms baseline models, showing that the objective function effectively minimizes counterfactual prediction errors. Following the setup from~\cite{jiang2022estimating}, we simulate outcomes by adjusting treatment flip rates (0.25, 0.5, 0.75, 1). As shown in Fig.~\ref{fig:4}, higher flip rates generally increase MSE, but CgNN maintains the lowest error.


\section{Conclusion}

\textbf{Summary of Contributions.} In this work, we propose CgNN, a novel method to address hidden confounder bias in network data while accounting for peer effects. CgNN distinguishes between ME, PE, and TE in networks. Since the underlying network structure captures critical information about hidden confounders, we design a GNN-driven IV approach that leverages the network structure as an IV to mitigate confounding bias. Combined with attention mechanisms, this approach distinguishes the varying influence of different peers, leading to more accurate effect estimation. Validated on two semi-synthetic datasets, CgNN demonstrates robustness in complex network settings.

\textbf{Limitations \& Future Work.} While CgNN effectively addresses hidden confounder bias in network data, it assumes the network structure provides sufficient information as valid IVs. Future work will focus on relaxing these assumptions to enhance its applicability.


\end{document}